\documentclass{article}

\usepackage[preprint]{neurips_2026}
\makeatletter
\renewcommand{\@notice}{}
\makeatother

\usepackage[utf8]{inputenc}
\usepackage[T1]{fontenc}
\usepackage{amsmath,amssymb,amsthm}
\usepackage{graphicx}
\usepackage{booktabs}
\usepackage{array}
\usepackage{makecell}
\usepackage{multirow}
\usepackage{xcolor}
\usepackage{colortbl}
\usepackage{algorithm}
\usepackage{algpseudocode}
\usepackage{microtype}
\usepackage{url}
\usepackage[bookmarks=true]{hyperref}
\usepackage{cleveref}
\usepackage{pifont}
\usepackage{fontawesome5}

\definecolor{vInk}{HTML}{16222E}
\definecolor{vBody}{HTML}{3A4A57}
\definecolor{vMuted}{HTML}{6B7A87}
\definecolor{vGreen}{HTML}{1D9F55}
\definecolor{vGreenF}{HTML}{E6F4EC}
\definecolor{vRed}{HTML}{BD2222}
\definecolor{vRedF}{HTML}{F8E7E7}
\definecolor{vBlue}{HTML}{1A6FD4}
\definecolor{vBlueF}{HTML}{E6EFFA}
\definecolor{vAmber}{HTML}{C77B13}
\definecolor{vAmberF}{HTML}{F9F0E5}
\definecolor{vLink}{HTML}{1D4E6F}

\newcommand{\papertitle}{Ratchet: How Reliable Must an LLM Judge Be to Retire a Skill?}

\hypersetup{
  colorlinks=true, linkcolor=vLink, citecolor=vLink, urlcolor=vLink,
  pdftitle={\papertitle},
  pdfauthor={Xing Zhang; Yanwei Cui; Guanghui Wang; Ziyuan Li; Wei Qiu; Bing Zhu; Peiyang He},
  pdfsubject={Skill-library maintenance for self-evolving LLM agents:
    the two-sided judge-reliability condition an eviction rule requires},
  pdfkeywords={self-evolving agents; skill libraries; library drift; LLM-as-judge;
    reward reliability; agent memory},
  bookmarksnumbered=true, bookmarksopen=true, bookmarksopenlevel=1,
  pdfdisplaydoctitle=true
}

\newtheorem{proposition}{Proposition}
\newcommand{\cmark}{\textcolor{vGreen}{\ding{51}}}
\newcommand{\xmark}{\textcolor{vMuted}{\ding{55}}}
\newcommand{\pmark}{\textcolor{vAmber}{$\circ$}}
\newcommand{\chat}{\hat{c}}
\newcommand{\rfp}{\rho_{\scriptscriptstyle F\to P}}
\newcommand{\rpf}{\rho_{\scriptscriptstyle P\to F}}
\newcommand{\Nmin}{N_{\min}}
\newcommand{\pt}{\pi_\tau}
\newcommand{\pstar}{p^{\star}}
\newcommand{\passk}{\textrm{pass@1}}
\newcommand{\bmbpp}{\textsc{MBPP+}/100}
\newcommand{\bswe}{\textsc{SWE-V}/150}
\newcommand{\brmain}{\textsc{Rpt}/71}
\newcommand{\brhard}{\textsc{Rpt}/133}
\newcommand{\brband}{\textsc{Rpt}/58}

\title{\papertitle}

\author{%
  Xing Zhang$^{1}$ \quad
  Yanwei Cui$^{1}$ \quad
  Guanghui Wang$^{1}$ \quad
  Ziyuan Li$^{2}$ \\
  \bfseries Wei Qiu$^{2}$ \quad
  Bing Zhu$^{2}$ \quad
  Peiyang He$^{1}$\thanks{Corresponding author: \texttt{peiyan@amazon.com}.}\\[2pt]
  \normalfont $^{1}$AWS Generative AI Innovation Center \\
  \normalfont $^{2}$HSBC Holdings Plc., HSBC Technology Center, China \\[4pt]
  \normalfont\small \href{https://github.com/amazon-science/Self-Evolving-Agents-Ratchet}%
  {\faGithub\; \nolinkurl{https://github.com/amazon-science/Self-Evolving-Agents-Ratchet}}
}

\begin{document}
\maketitle

\begin{abstract}
A large language model (LLM) agent that writes and edits its own skill library must also
decide which skills to keep, from one noisy scalar per skill. The answer is exact: a judge
scoring failures as passes at rate $(1-\tau)/2$ or above retires nothing, at any sample
size, for eviction margin $\tau$. Audits find that machinery is rarely
built: LLM-written skills are worth $+0.0$ percentage points (pp) against a no-skill control,
human-written ones $+16.2$pp. Unmaintained, a library enters \emph{library drift}, growing
until injecting a skill scores worse than injecting nothing. \textbf{Ratchet} repairs this: it
evicts each skill on its measured contribution, caps the library at width $C$, and constrains
synthesis, lifting held-out $\passk$ by $+0.328$ on a hard MBPP+ slice.
The matching non-divergence bound is finite for exactly two reasons, $C$ and $\tau$.
Our contribution is the condition this repair carries and no deployed system states. In
reference-free domains the scalar comes from an LLM judge, whose two error directions,
modelled as a binary channel, behave nothing alike. Passes scored as failures cost
sample efficiency, which more trials buy back; failures scored as passes displace the eviction
statistic, and no correction inside the rule recovers it. End-task score is a poor alarm,
moving by at most a fifth of the governed lift and not monotonically in the rate. We prove
both edges of the certifiable region, confirm them in a running loop, and place a judge on a
known side in one offline pass.
\end{abstract}

\section{Introduction}
\label{sec:intro}

A skill library lets a frozen LLM keep something from every task it solves, and
Voyager~\citep{wang2023voyager} set the pattern: write the successful procedure down, retrieve it when a
similar task arrives, compound without a gradient step. That promise has now been audited and it did not
hold. SkillsBench~\citep{li2026skillsbench} scores LLM-authored skills at $+0.0$pp over a no-skill control
on tasks where human-authored skills score $+16.2$pp, and a survey of over twenty such
systems~\citep{zhang2026ecs} finds the machinery for versioning, conflict detection, and deprecation
``largely neglected.'' The libraries compound; the agents do not. We argue the shortfall is a missing
maintenance rule rather than poor authorship, and that supplying one carries a reliability condition on the
reward it reads that no deployed system states.

Name the missing rule precisely and both halves of the result follow. These systems author, retrieve, and
inject skills, but none estimates what an individual skill is worth and acts on the estimate, so a harmful
artifact is never removed. Add that estimate and the removal rule it feeds, and read what results as a
\emph{feedback loop}: the library is the plant, the maintenance rule the controller, and the measured
outcome of using a skill the only sensor. That reading makes the repair and its reliability condition one
statement, and it predicts where the repair breaks.

\paragraph{The failure has a definition, not just a name.}
\emph{Library drift} is testable rather than rhetorical: a library is harmful, not
merely unhelpful, once expected held-out $\passk$ with it falls below expected $\passk$ with it withheld, a
control the same model supplies (\Cref{eq:harm}). Nothing throws, and an adaptive router masks it by declining more often as the
library degrades. Prior work names the symptoms, retrieval dilution and stagnation, without a definition a
run can be tested against; this one costs a single round with injection switched off.

\paragraph{The repair works, and its bound has content in which quantities appear.}
\textbf{Ratchet} is the controller. It holds the author fixed, one frozen model in every role, and reads one
scalar per skill. Three levers act on that scalar: evict a skill once its measured contribution reaches
$-\tau$ on at least $\Nmin$ trials, hold at most $C$ skills active and evict the weakest when synthesis would
overflow, and constrain new writing with one
authoring prior. On a hard MBPP+ slice this takes held-out $\passk$ from $0.258$ to a $0.658$ peak, a rolling
gain of $+0.328$. \Cref{prop:floor} bounds how far below the no-skill control a governed library can fall,
and its content is not the numeric value but which quantities appear in it: the width $C$ and the margin
$\tau$, both finite by construction. Remove either and the statement has nothing left to say, which is the
sense in which running the loop open is unbounded.

\begin{figure}[t]
  \centering
  \includegraphics[width=\textwidth]{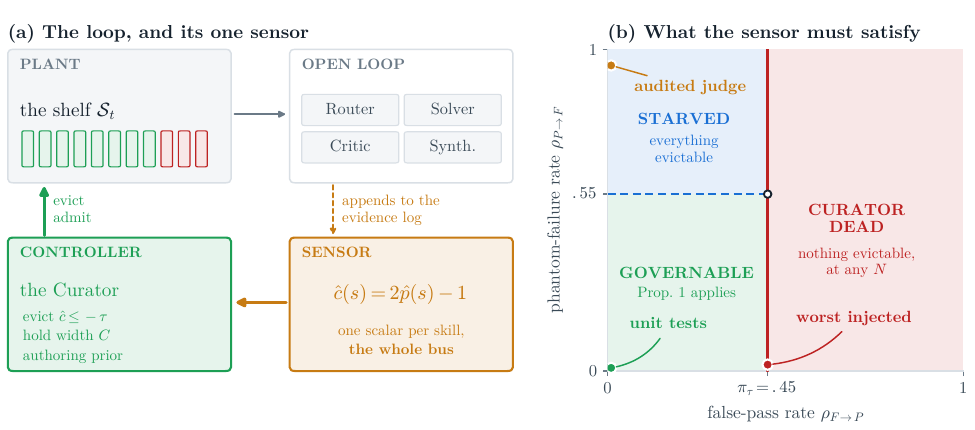}
  \caption{\textbf{The loop, and the specification its sensor must meet.} \emph{(a)} One frozen model
  fills five roles over one append-only log. Four grow the library with no maintenance; the fifth, the
  Curator, closes it with three levers (green), reading exactly one quantity, the measured contribution $\chat(s)$
  supplied by the sensor (amber). \emph{(b)} The certification plane. Eviction requires a true pass rate
  below $(\pt - \rfp)/\kappa$, and the two exact inequalities that bound it are both axis-parallel, so the
  governable region is a rectangle whose corner sits at $(\pt, 1-\pt)$: past $\rfp = \pt = (1-\tau)/2$
  nothing is evictable at any sample size, and past $\rpf = 1-\pt$ everything is. A unit-test grader sits
  at the origin, while the audited judge of \Cref{sec:reward-gate} clears the false-pass threshold by a
  wide margin but lies past the starvation edge, safe from the fatal direction and paying for it in
  resolution.}
  \label{fig:thesis}
\end{figure}

\paragraph{The sensor specification, and what it costs to violate it.}
\Cref{prop:floor} needs the contribution estimate to be unbiased, which unit tests supply for free and the
deployments this machinery is built for do not: they are reference-free, they score with an LLM judge, and
judge errors are structured rather than stochastic~\citep{wang2024large,stureborg2024large}. Modelling the
judge as a binary channel splits those errors into two directions that behave nothing alike. A true pass
scored as a failure costs evidence, and evidence is purchasable with more trials. A true failure scored as a
pass moves the eviction statistic in the one direction that prevents eviction, by the largest amount for
exactly the skills that most need evicting, and once that rate reaches $\pt = (1-\tau)/2$ nothing is evicted
at any sample size (\Cref{prop:floorbias}). More data is no remedy for a displaced mean, and little on an
end-task dashboard says so, because a disabled controller changes which artifacts survive long before it
changes any pass rate. Nor is the harm ordered by the rate: it peaks \emph{at} the boundary, so the most
dangerous judge is the half-blind one rather than the blindest. Two exact inequalities therefore partition
every reward into a governable region, one where the controller is dead, and one where it evicts
indiscriminately (\Cref{fig:thesis}b), and one offline pass places a deployment on that plane.

\paragraph{What is new.}
The derivation, and what it exposes. Library drift gets a definition rather than a name, testable in one
round. The levers are derived rather than proposed, as exactly the two quantities a floor
needs to be finite, so the absence of any bound for an unmaintained library is a consequence rather than an
assertion. The mechanisms themselves we take as given, concurrent procedural-memory systems having reached
score-based maintenance online, one of them under a bounded capacity (\Cref{sec:problem}). And the reliability
that rule inherits from its sensor becomes a specification with three properties. It is a \emph{single}
inequality, \Cref{prop:floorbias} returning \Cref{prop:floor} at zero error, so the guarantee and its
precondition cannot be read apart. It is \emph{two-sided}, the second edge being what places the audited judge
outside the governable region despite its clearing the false-pass cliff by a factor of $45$. And it is a
\emph{precondition}, checkable before a maintenance rule is switched on rather than revealed by a failure
after.\footnote{Earlier reports of ours cover the diagnosis~\citep{zhang2026librarydrift} and the judge
channel~\citep{zhang2026blindcurator} separately; this version integrates and supersedes both, and stands on
its own.}

\paragraph{Where this sits relative to noisy-verifier learning.}
A concurrent literature asks a similar question of policy optimisation and reaches a reassuring answer, which
is the sharpest way to locate this result. \citet{cai2025noisyverifier} model an imperfect verifier as a
reward channel with two asymmetric rates, the same abstraction we use, and derive corrections restoring an
unbiased policy gradient; \citet{plesner2026imperfect} measure reinforcement learning from verifiable rewards
tolerating noise up to $15\%$ at within two points of clean validation accuracy. Both study a \emph{gradient},
a consumer that averages many rewards, where bias is a term to correct or to absorb. An eviction rule is a
different consumer. It thresholds one displaced mean per artifact, so the bias does not average away, and past
$\pt$ no correction available inside the rule recovers it. Closest in structure,
VRR-Stop~\citep{wu2026vrrstop} separates verifier false acceptance from false rejection in a verify-repair
loop and reports acceptance rising while true validity falls, the same silence we measure at a different
decision: theirs is when to stop repairing one plan, ours which artifacts a library retires across rounds.
What none of these supplies, and \Cref{sec:reward-gate} does, is a rate measurable before the loop is
switched on.

\section{Library drift: when a skill library becomes harmful, and why it is rarely measured}
\label{sec:problem}

\paragraph{Library drift, defined against a control the model itself supplies.}
Let $\mathcal{S}_t$ be the skill set an agent carries into round $t$, and $p_0$ the pass rate of
the same frozen model on the same task distribution with no skill injected. Library drift is then exactly
this: the library is harmful when
\begin{equation}
  \mathbb{E}\!\left[\passk \mid \mathcal{S}_t\right] \;<\; \mathbb{E}[p_0]
  \qquad\text{for some } t > 0.
  \label{eq:harm}
\end{equation}
Three properties earn this definition its place. It is a statement about the set, not any member, since every
skill can look defensible in isolation while \Cref{eq:harm} holds; the comparison is within-model, so no
capability difference can produce it; and the control costs one round with injection switched off, which makes
it operational. What produces it is retrieval dilution: a growing \emph{shelf} returns near-duplicates and
superseded material at constant recall, and a superseded skill in the prompt is worse than an absent one,
arguing for a specific wrong approach instead of declining to help. No stage raises an error, so the agent
simply fails more often, which from outside looks like a hard task distribution.

Two properties make this a distinct failure mode rather than a restatement of a known one, and both fix the
shape of the cure. It requires \emph{persistent cross-task} artifacts, which is what puts within-episode
correction out of reach: Reflexion, Self-Refine, and ReAct discard the critique at the episode
boundary~\citep{shinn2023reflexion,madaan2023selfrefine,yao2023react}, and
Toolformer~\citep{schick2023toolformer} learns tool use with no library to maintain. And it acts through
retrieval rather than representation, which is what separates it from catastrophic forgetting, the closest
established phenomenon: there new gradients overwrite representations and regularisers such as
EWC~\citep{kirkpatrick2017ewc} penalise movement in parameter space, while here no parameter moves. So the
cure is eviction under a bounded width, not a penalty term and not a better critic. Library drift begins
where artifacts outlive their episode.

\paragraph{Why an aggregate curve will not show it, and what will.}
Held-out $\passk$ folds task difficulty, sampling variance, and library quality into one scalar, and an
adaptive router defends it: declining when nothing applies is the router's job, so the aggregate holds up on
progressive abstention while the shelf hollows out. What it lacks even once it moves is resolution, since it
cannot say which skill is responsible and so cannot drive a removal. Because every stage appends to one
evidence log, a per-skill quantity is available with no new instrumentation: the \emph{measured
contribution} $\chat(s)$ of \Cref{eq:chat}, attributable to a single artifact by construction and the entire
input to the rule of \Cref{sec:loop}. That quantity is what the rest of this paper is about, first as what
repairs library drift and then as what a fallible judge corrupts.

\paragraph{Cross-task libraries invest in authorship.}
Voyager~\citep{wang2023voyager} grows an executable library, ExpeL~\citep{zhao2024expel} textual insights,
Agent Workflow Memory~\citep{wang2024awm} reusable workflows induced from experience, and
AutoManual~\citep{chen2024automanual} a flat per-domain manual, none of them retiring an entry on its
measured outcome. Adjacent machinery is missing a different piece: prompt
optimisers~\citep{khattab2023dspy,yang2023opro,yuksekgonul2024textgrad} carry no per-skill evidence, memory
hierarchies~\citep{packer2023memgpt,park2023generative} no typed retirable artifact, and
retrieval-augmented generation~\citep{lewis2020rag} conditions on a store it never prunes on outcome. Recent
work extends authorship further still, into strategic principles~\citep{wu2025evolver}, trace-induced
skills~\citep{alibaba2026trace2skill}, meta-skills~\citep{huang2025cascade}, layered
decomposition~\citep{liang2026ssl}, failure history as metadata~\citep{wang2026strategygenes}, and libraries
coupled to weight updates~\citep{xia2026skillrl}. Every one of them writes; none retires on measured
contribution.

\paragraph{The few that maintain, and the reliability question none of them asks.}
AutoSkill supplies the closest machinery, versioning plus a standardised schema, but no eviction on measured
per-task contribution~\citep{yang2026autoskill}, while SkillsVote~\citep{liu2026skillsvote}, Dynamic Skill
Lifecycle Management~\citep{shen2026dynamic}, SkillAdaptor~\citep{yu2026skilladaptor},
and insight governance for verbal reinforcement learning~\citep{cui2026feedback} curate on self-report or
recency rather than on audited contribution under a bounded width. SkillAudit~\citep{gao2026skillaudit} attacks
the signal problem from the opposite side, removing the privileged-feedback assumption entirely by executing
each task with and without a candidate skill and reading the divergence, which supplies a contribution
estimate without a grader but says nothing about how reliable an estimate a rule needs. Closest and concurrent,
ProcMEM~\citep{mi2026procmem} pairs a fixed-capacity pool with
score-based pruning, and MACLA~\citep{forouzandeh2025macla} prunes on Bayesian per-procedure reliability
without bounding capacity, both learned online; we reach ProcMEM's pairing from the bound of
\Cref{sec:theory} instead. What none of them states is a
condition on the reliability of the signal the maintenance runs on, and that is where a second literature
attaches. Strong judges agree with human preference above $80\%$ of the time~\citep{zheng2023llmjudge}, but
their errors are systematic rather than
stochastic~\citep{stureborg2024large,wang2024large,li2024llms,gurram2026agentprop}, and noisy-label theory
separates symmetric from class-conditional noise~\citep{natarajan2013learning}. Carried to a lifecycle rather
than a classifier, that distinction sharpens into a specification: the \emph{direction} of the asymmetry, not
the error rate, decides whether the controller keeps working (\Cref{sec:reward}).

\paragraph{What the field has built, disaggregated.}
The claim that authorship is well served and maintenance is not can be checked feature by feature rather than
asserted, and \Cref{tab:capabilities} does that across sixteen systems and eight design axes. Read it as
\emph{feature presence}, not as head-to-head performance, which it cannot establish and which the ablations of
\Cref{sec:evidence-ablations} establish here instead. The shape it shows is the argument: the authoring
columns are largely filled, the governance block nearly empty, its rightmost column carried by one concurrent
system. That empty block is the object of this paper, and that column is what \Cref{sec:theory} shows a floor
cannot do without.

\newcommand{\colhead}[1]{\makecell{\footnotesize #1}}
\begin{table}[t]
  \centering
  \footnotesize
  \setlength{\extrarowheight}{1pt}
  \caption{\textbf{Design-axis presence across skill-library systems}
  (\cmark\,present, \xmark\,absent, \pmark\,partial). Columns group into \emph{signal quality},
  \emph{authoring}, and \emph{governance}; the shaded \textcolor{vGreen}{governance} block is where prior
  systems are nearly always empty and where the ablations of \Cref{sec:evidence-ablations} locate the
  load-bearing mechanisms. This is a presence table, not a benchmark. ${}^{\dagger}$SkillRL updates weights,
  a different regime, and is included for contrast.}
  \label{tab:capabilities}
  \renewcommand{\arraystretch}{1.2}
  \setlength{\tabcolsep}{5pt}
  \newcolumntype{G}{>{\columncolor{vGreenF}}c}
  \resizebox{\textwidth}{!}{%
  \begin{tabular}{l cc ccc GGG}
    \toprule
    & \multicolumn{2}{c}{\emph{signal quality}} & \multicolumn{3}{c}{\emph{authoring}} & \multicolumn{3}{c}{\emph{\textcolor{vGreen}{governance}}} \\
    \cmidrule(lr){2-3} \cmidrule(lr){4-6} \cmidrule(lr){7-9}
    System
      & \colhead{Separate\\Critic}
      & \colhead{Evidence\\log}
      & \colhead{Failure-clust.\\synthesis}
      & \colhead{Typed skill\\schema}
      & \colhead{Meta-skill\\layer}
      & \colhead{Pattern\\canonicalis.}
      & \colhead{Eviction on\\contribution}
      & \colhead{Bounded\\width $C$} \\
    \midrule
    MemGPT~\citep{packer2023memgpt}            & \xmark & \pmark & N/A       & \pmark & \xmark & \xmark & \pmark & \pmark \\
    DSPy~\citep{khattab2023dspy}               & \pmark & \pmark & N/A       & \cmark & \pmark & \xmark & \xmark & \xmark \\
    Voyager~\citep{wang2023voyager}            & \xmark & \pmark & \pmark & \cmark & \pmark & \xmark & \xmark & \xmark \\
    AWM~\citep{wang2024awm}                    & \xmark & \pmark & \xmark & \cmark & \xmark & \xmark & \xmark & \xmark \\
    AutoManual~\citep{chen2024automanual}      & \xmark & \pmark & \pmark & \pmark & \xmark & \xmark & \xmark & N/A       \\
    ExpeL~\citep{zhao2024expel}                & \pmark & \cmark & \pmark & \xmark & \xmark & \xmark & \pmark & \xmark \\
    CASCADE~\citep{huang2025cascade}           & \pmark & \pmark & \xmark & \pmark & \cmark & \xmark & \xmark & \xmark \\
    EvolveR~\citep{wu2025evolver}              & \pmark & \pmark & \pmark & \pmark & \xmark & \pmark & \pmark & \xmark \\
    SSL~\citep{liang2026ssl}                   & \xmark & \xmark & \xmark & \cmark & \pmark & \xmark & \xmark & \xmark \\
    AutoSkill~\citep{yang2026autoskill}        & \xmark & \pmark & \xmark & \cmark & \xmark & \xmark & \pmark & \xmark \\
    Strategy~Genes~\citep{wang2026strategygenes} & \xmark & \pmark & \pmark & \cmark & \xmark & \xmark & \xmark & \pmark \\
    SkillRL~\citep{xia2026skillrl}${}^{\dagger}$ & \pmark & \pmark & \pmark & \pmark & \pmark & \xmark & \xmark & \xmark \\
    Trace2Skill~\citep{alibaba2026trace2skill} & \pmark & \cmark & \pmark & \pmark & \xmark & \pmark & \xmark & \xmark \\
    SkillAudit~\citep{gao2026skillaudit}       & \cmark & \pmark & \xmark & \pmark & \xmark & \xmark & \pmark & \xmark \\
    MACLA~\citep{forouzandeh2025macla}         & \xmark & \pmark & \pmark & \pmark & \xmark & \xmark & \pmark & \xmark \\
    ProcMEM~\citep{mi2026procmem}              & \pmark & \xmark & \pmark & \cmark & \xmark & \pmark & \cmark & \cmark \\
    \midrule
    \rowcolor{vGreenF}
    \textbf{Ratchet (this paper)}              & \cmark & \cmark & \cmark & \cmark & \cmark & \cmark & \cmark & \cmark \\
    \bottomrule
  \end{tabular}%
  }
\end{table}

\section{Ratchet: five roles, one log, three levers}
\label{sec:loop}

The design premise is austerity: hold the plant and the four open-loop roles fixed, a frozen LLM with no
weight update, and add only a controller. The name records what that controller is for, a mechanism that lets
the library advance and resists its sliding back. Better authorship is not worthless, but given an adequate
author, control binds. Austerity is also what makes the measurements interpretable, since a fixed author
charges any measured change to the controller, and what makes the corruption analysis of \Cref{sec:reward}
possible at all: a controller reading two scalars can be attacked by corrupting two scalars.

\paragraph{Five roles over one durable object.}
One frozen model is invoked with five role prompts against a shared append-only \emph{evidence log}. Four
run the plant open-loop. The \textbf{Router} sees a task and the shelf and returns at most one skill, or
declines; the one-skill restriction keeps attribution unambiguous, since every outcome is charged either to
a named skill or to the bare model. The \textbf{Solver} attempts the task with that skill injected as
guidance and is graded. The \textbf{Critic} reads each failure with its task, routed skill, output, and
grader trace, and returns an attribution label plus a short pattern string. The \textbf{Synthesizer}
clusters recent patterns and writes new skills under the authoring prior. The fifth, the
\textbf{Curator}, is the controller and holds the three levers below (\Cref{app:config} gives all five
prompts). The log is the only durable state, holding \emph{capsules}, one per (task, skill, attempt), and
\emph{verdicts}, one per failure capsule. Nothing is deleted: eviction flips a status flag from
\textsc{active} to \textsc{deprecated}, dropping the skill from retrieval while its content and evidence
stay addressable, so every control decision is reconstructible from the log alone, which lets
\Cref{sec:reward} re-read it under a corrupted sensor with nothing else changed.

\paragraph{Lever 1: evict on measured contribution.}
The Curator computes, for each active skill,
\begin{equation}
  \chat(s) \;:=\; \frac{\#\text{pass}(s) - \#\text{fail}(s)}{n(s)}
  \;=\; 2\hat{p}(s) - 1,
  \label{eq:chat}
\end{equation}
the mean signed outcome over the $n(s)$ trials the Router sent to $s$. This single scalar is the entire
sensor reading, which is why \Cref{sec:reward} can attack the loop through it alone. Eviction requires two
conditions jointly, $n(s) \ge \Nmin$ and $\chat(s) \le -\tau$, at defaults $\tau=0.10$, $\Nmin=100$.

The two conditions do different jobs and both are load-bearing. The threshold $\tau$ decides \emph{which}
skills are candidates; the evidence floor $\Nmin$ decides \emph{when} the estimate may be acted on, and it is
not a detail, \Cref{sec:evidence} measuring a low floor with a zero threshold driving the loop below the
no-skill control, worse than no maintenance at all. Firing on the conjunction is what separates eviction from
the recency and self-report heuristics of \Cref{sec:problem}, since a skill goes only once the log has the
trials to say it is harmful rather than unlucky. Note what $\chat$ estimates: trials accumulate only on the
subpopulation the Router chose to send, so it is routed-conditional rather than a full-distribution contrast.
That is deliberate, being the target a removal decision needs, and \Cref{sec:theory} bounds that same target
rather than a stronger one it could not support; the cost is a survivorship confound (\Cref{sec:close}).

\paragraph{Lever 2: hold the shelf to a fixed width.}
At most $C$ skills are \textsc{active}, default $C=50$. When synthesis would exceed $C$, the Curator evicts
$\arg\min_s \chat(s)$. Skills compete for a fixed-width shelf, so retrieval precision does not decay as the
library grows, and $C$ is one of the two finite quantities making \Cref{prop:floor} say anything. The two
levers are not redundant. Lever 1 fires only on skills that have earned $\Nmin$ trials, so it says nothing
about a young skill; lever 2 fires regardless of evidence, the cruder instrument but the one that binds the
set the union bound of \Cref{prop:floor} runs over. Width is therefore what keeps the floor finite while
contribution-based eviction is what would let the shelf rise, and \Cref{sec:reward} finds them separating
exactly that way under a corrupted sensor.

\paragraph{Lever 3: constrain what may be written.}
A single short \emph{meta-skill} document per suite, carrying a schema lock and authoring guidance, enters
the Synthesizer's prompt, pushing new skills toward reusable patterns, each with an applicability
condition, a key insight, known pitfalls, and a post-condition check, and away from task-specific snippets.
It has a second effect the ablations surfaced: the stylistic homogeneity it imposes makes embedding-based
deduplication redundant at this scale (\Cref{sec:evidence}). Two supporting mechanisms complete the loop.
Before clustering, near-duplicate pattern labels are collapsed by union-find over embedding cosine similarity
at $0.85$, and a \emph{cover-guard} skips any cluster an active skill already covers. Separately, a round whose
held-out $\passk$ falls more than $\tau_{\mathrm{rb}}=0.10$ below the running best is flagged, and the best
snapshot restored after five \emph{consecutive} flags; that net fired twice in 300 rounds.

\paragraph{One round, what it costs, and what it gives up.}
\Cref{alg:round} assembles the roles; every state change is either an append to the log or a status-flag
flip. The \textsc{Govern} block is the whole controller, and its interface to the plant is two scalars per
skill, $\chat(s)$ and $n(s)$: no model call, no text, no metadata. That narrowness is what makes
\Cref{sec:theory} tractable and \Cref{sec:reward} dangerous, since corrupting one quantity disables the
block entirely. It also makes the controller free, since per-round cost is one Router and one Solver call per
task, one Critic call per failure, and one Synthesizer call per uncovered cluster, all of which the open loop
already pays; governance is pure arithmetic overhead, and nothing updates a weight, so the method needs no
accelerator of its own (\Cref{app:config}). What that austerity costs is composition and text: no task ever
receives two skills, and the Curator cannot recognise two skills as near-duplicates except through the
outcomes they produce, which is why the authoring prior rather than the Curator carries deduplication.

\begin{algorithm}[t]
\caption{One governed round. Frozen model $M$ in all five roles; log $\mathcal{L}$ append-only;
shelf $\mathcal{S}$; the only \emph{quantities} the Curator reads are $\chat(s)$ and $n(s)$, both
recomputed from $\mathcal{L}$.}
\label{alg:round}
\small
\begin{algorithmic}[1]
\Require shelf $\mathcal{S}$ ($|\mathcal{S}| \le C$), meta-skill $m$, log $\mathcal{L}$,
params $(\tau, \Nmin, C, \tau_{\mathrm{rb}})$
\Statex \hspace{-1.2em}\textcolor{vBody}{\textsc{Act} \; $\cdot$ \; \emph{four roles, no state change
beyond appends}}
\State \textbf{for} $x \in \mathcal{X}_{\mathrm{eval}}$ \textbf{then} $\mathcal{X}_{\mathrm{train}}$:
  \; $s \gets M_{\textsc{route}}(x, \mathcal{S})$ \Comment{one \textsc{active} skill, or decline}
\State \hskip 1.1em $\mathcal{L} \gets \mathcal{L} \,\Vert\, \textsc{capsule}\bigl(x, s,
  M_{\textsc{solve}}(x, s)\bigr)$ \Comment{graded outcome}
\State \textbf{for} each train-failure capsule $c$: \;
  $\mathcal{L} \gets \mathcal{L} \,\Vert\, \textsc{verdict}\bigl(M_{\textsc{critic}}(c)\bigr)$
  \Comment{label $+$ pattern string}
\Statex \hspace{-1.2em}\textcolor{vAmber}{\textsc{Grow} \; $\cdot$ \; \emph{failures become
candidate skills}}
\State $K \gets \textsc{cluster}\bigl(\textsc{canonicalise}(\text{recent patterns})\bigr)$
\State \textbf{for} $k \in K$, $|k| \ge 3$, $k$ uncovered: \;
  $\mathcal{S} \gets \mathcal{S} \cup \{ M_{\textsc{synth}}(k, m) \}$
  \Comment{prior $m$ constrains the write}
\Statex \hspace{-1.2em}\textcolor{vGreen}{\textsc{Govern} \; $\cdot$ \; \emph{levers 1 and 2;
this is the entire cure}}
\State \textbf{for} $s \in \mathcal{S}$ with $n(s) \ge \Nmin$: \;
  \textbf{if} $\chat(s) \le -\tau$ \textbf{then} flag $s$ \textsc{deprecated}
  \Comment{\emph{lever 1}: floor, then threshold}
\State \textbf{while} $|\mathcal{S}| > C$: \; evict $\arg\min_{s \in \mathcal{S}} \chat(s)$
  \Comment{\emph{lever 2}: fixed-width shelf}
\Statex \hspace{-1.2em}\textcolor{vMuted}{\textsc{Guard} \; $\cdot$ \; \emph{fired twice in
300 round-decisions}}
\State \textbf{if} held-out $\passk < \mathrm{best} - \tau_{\mathrm{rb}}$ for $5$
  consecutive rounds \textbf{then} restore best $\mathcal{S}$
\State \Return $\mathcal{S}, \mathcal{L}$
\end{algorithmic}
\end{algorithm}

\section{Why a closed loop has a floor, and what buys the finiteness}
\label{sec:theory}

Measurements cannot say whether anything \emph{prevents} a governed shelf from sliding below its
no-skill control or whether it merely happens not to. Two propositions answer that as one statement:
\Cref{prop:floor} exhibits a floor whose finiteness comes from $C$ and $\tau$ alone, and
\Cref{prop:floorbias} generalises it to a fallible sensor and exposes the boundary \Cref{sec:reward}
measures.

\paragraph{Setup.}
Fix a task distribution $\mathcal{D}$. For a task $x$, let $p_0(x)$ and $p(x \mid s)$ be the pass
probabilities of the frozen model with no skill and with $s$ injected, and let $\mathcal{D}_s$ be the
distribution of tasks the Router sends to $s$. Define the contribution on that subpopulation,
\begin{equation}
  c(s) \;:=\; \mathbb{E}_{x\sim\mathcal{D}_s}\!\left[\, p(x \mid s) - p_0(x) \,\right].
  \label{eq:contrib}
\end{equation}
This is both what the log estimates through \Cref{eq:chat} and what governs outcomes on the tasks $s$
receives; defining it on $\mathcal{D}_s$ removes any separate coverage assumption, the accumulating trials
being draws from $\mathcal{D}_s$ by construction. Eviction fires when $n(s) \ge \Nmin$ and
$\chat(s) \le -\tau$, the rule of \Cref{sec:loop}.

\begin{proposition}[Governed shelves have a floor]
\label{prop:floor}
Suppose (i) the Router returns either a decline or a member of $\mathcal{S}_t$ for every task, and
declining yields $p_0(x)$; (ii) $\chat(s)$ is unbiased and consistent for $c(s)$ as defined in
\Cref{eq:contrib}; and (iii) the evidence floor is set so that a concentration inequality gives a
uniform accuracy guarantee, namely that after $\Nmin$ trials $|\chat(s) - c(s)| \le \epsilon$ holds
simultaneously for all $s \in \mathcal{S}_t$ with probability at least $1 - C\delta$, each skill
contributing at most $\delta$ to the failure probability. Then
\begin{equation}
  \mathbb{E}\!\left[\passk\right] \;\ge\; \mathbb{E}[p_0] \;-\; (\tau + \epsilon)
  \;-\; C\,\delta.
  \label{eq:floor}
\end{equation}
\end{proposition}

In one line: a skill that survives failed the eviction test, so its true contribution is at least
$-\tau - \epsilon$; averaging that over the routing distribution and charging the concentration event its
$C\delta$ gives \Cref{eq:floor} (\Cref{app:proofs}).

\paragraph{What to take from the floor, and what not to.}
At the defaults ($\tau=0.10$, $\Nmin=100$, $C=50$, $\delta=10^{-3}$ per skill) a Hoeffding radius gives
$\epsilon \approx 0.39$ and $C\delta = 0.05$, so \Cref{eq:floor} reads $\mathbb{E}[p_0] - 0.30$ on the
pass-rate scale (\Cref{app:proofs}). That is loose, and \Cref{sec:evidence}'s $+0.328$ shows how loose. Two
structural facts matter more. The bound constrains the downside only, never that the shelf will rise, so it
complements the measurements rather than substituting for them. And its right-hand side is finite for one
reason, that $C$ and $\tau$ are: drop the width and the union bound runs over an unbounded set, so $C\delta$
diverges; drop the threshold and survivors carry no lower bound on $c(s)$. An open loop therefore has no
counterpart to \Cref{eq:floor}~\citep{wang2023voyager,zhao2024expel,chen2024automanual}, which is the formal
sense in which library drift is a control failure and not an authorship one.

\subsection{The assumption, and what happens when it fails}
\label{sec:theory-bias}

Assumption (ii) is the sensor specification, stated. With a deterministic grader it is free: unit tests
return the outcome, so $\chat$ averages truths. In reference-free domains the reward is an LLM judge and
its errors are structured. Where potential-based reward shaping asks which reward transformations preserve
the optimal policy~\citep{ng1999pbrs}, the lifecycle question is dual: which preserve an \emph{eviction
rule}. With $y \in \{0,1\}$ the true trial outcome and $\tilde{y}$ the scored one, a two-parameter channel
answers it,
\begin{equation}
  \Pr[\tilde{y}{=}1 \mid y{=}0] = \rfp, \qquad
  \Pr[\tilde{y}{=}0 \mid y{=}1] = \rpf, \qquad
  \kappa := 1 - \rfp - \rpf > 0,
  \label{eq:channel}
\end{equation}
subscripts reading true$\,\to\,$scored. We avoid the false-positive and false-negative labels, which swap
meaning with the choice of positive class, and the two directions are not interchangeable: $\rfp$ hides a
failure and $\rpf$ invents one. This channel is not new and its recurrence is the point:
\citet{cai2025noisyverifier} write the same two rates for an imperfect verifier feeding a policy gradient, and
their remedy, an importance-weighted backward correction, is available precisely because a gradient consumes
rewards in expectation, so an unbiased surrogate reward restores an unbiased estimator. The consumer here
admits no such move: the eviction rule reads one mean per artifact against a fixed threshold, and correcting
the reward in expectation does not relocate a threshold the displacement has already carried out of the
domain. Worse, observed failures are the loop's only substrate, feeding synthesis as well as eviction, so
$\rfp$ deletes that substrate at the source. A skill whose true pass rate on its routed subpopulation is
$\bar{p}(s)$ has a scored pass rate concentrating on
\begin{equation}
  p_{\mathrm{sc}}(s) \;=\; \kappa\,\bar{p}(s) + \rfp.
  \label{eq:pobs}
\end{equation}
The eviction test $\chat(s) \le -\tau$ is the test $\hat{p}(s) \le \pt := (1-\tau)/2$, so everything
follows from how \Cref{eq:pobs} moves the scored rate relative to a fixed threshold, which it does in two
ways that have nothing in common (\Cref{fig:frontier}a).

\emph{Symmetric noise rescales, and the ordering survives.} At $\rfp = \rpf = \rho < \tfrac12$,
\Cref{eq:pobs} becomes $p_{\mathrm{sc}} - \tfrac12 = (1-2\rho)(\bar{p} - \tfrac12)$, a contraction toward
$\tfrac12$, and contraction about a fixed point preserves order, so a genuinely harmful skill still crosses,
at an inflated effective threshold $\tau_{\mathrm{eff}} = \tau/(1-2\rho)$; and since the estimator must
resolve means separated by $(1-2\rho)$, the evidence floor grows as $\Nmin \propto (1-2\rho)^{-2}$. Both
costs are finite for every $\rho < \tfrac12$, so the loop pays in evidence, and evidence is purchasable.

\emph{False-pass bias translates, and the specification is two-sided.} At $\rpf = 0$, $\rfp > 0$,
\Cref{eq:pobs} gives $p_{\mathrm{sc}} = \bar{p} + \rfp(1 - \bar{p})$, an upward displacement largest for
the smallest $\bar p$, which is to say for exactly the skills eviction exists to remove. In general eviction
requires $\bar{p}(s) \le \pstar := (\pt - \rfp)/\kappa$, and that test informs only while
$\pstar \in (0,1)$, a rule firing for no skill and one firing for every skill being equally uninformative.
Both ends are exact and each involves one rate alone: $\pstar \le 0$ iff $\rfp \ge \pt = (1-\tau)/2$, and
$\pstar \ge 1$ iff $\rpf \ge 1 - \pt$. The certifiable set is therefore the rectangle
$[0,\pt) \times [0, 1-\pt)$ with corner $(\pt, 1-\pt)$ (\Cref{fig:thesis}b), and its edges fail
differently. Past $\pt$ the condition is unsatisfiable even by a skill that never solves anything, and
sample size does not enter, more trials only concentrating the estimator around a displaced mean; past
$1-\pt$ the controller instead acts on everything, forfeiting the resolution eviction exists to supply. The
second edge binds in practice, because a judge tuned away from the fatal one lands on it
(\Cref{sec:reward-gate}). Both predictions are boundaries, not slopes (\Cref{sec:reward-frontier}).

\begin{proposition}[The floor under a fallible reward]
\label{prop:floorbias}
Assume the Router condition of \Cref{prop:floor} and the channel of \Cref{eq:channel} with known
bounds $\rfp \le \bar\rho_{F}$, $\rpf \le \bar\rho_{P}$, and $\kappa \ge \underline\kappa > 0$.
Choose $\Nmin$ so that every active skill's scored pass rate lies within $\epsilon$ of its mean with
probability at least $1-\delta$. Then
\begin{equation}
  \mathbb{E}\!\left[\passk\right] \;\ge\;
  \mathbb{E}[p_0]
  \;-\; \frac{\tau/2 + \epsilon + \bar\rho_{F}}{\underline\kappa}
  \;-\; \tfrac{1}{2}\bigl(1 - \underline\kappa\bigr)
  \;-\; C\,\delta.
  \label{eq:floorbias}
\end{equation}
\end{proposition}

The proof is \Cref{prop:floor}'s with one step added: \Cref{eq:pobs} is strictly increasing in $\bar p$, so
a survivor's scored rate inverts to a bound on its true rate, $\bar{p}(s) \ge
(\pt - \epsilon - \bar\rho_{F})/\underline\kappa$, and that inversion is where $\underline\kappa$ enters as a
divisor (\Cref{app:proofs}). The divisor is the whole story: it is what makes the guarantee fail
discontinuously rather than gracefully.

\paragraph{The two propositions are one statement.}
Set $\bar\rho_{F} = \bar\rho_{P} = 0$. Then $\underline\kappa = 1$, both corruption terms vanish, and
\Cref{eq:floorbias} reduces to \Cref{eq:floor}, differing only in convention: \Cref{prop:floor} reads the
margin on the $[-1,1]$ contribution scale and \Cref{prop:floorbias} on the $[0,1]$ pass-rate scale of the
implemented statistic, the two related by $\chat = 2\hat p - 1$. The same inequality, at an honest reward, is
a guarantee; at a biased one, a warning; and at $\rfp = \pt$, vacuous. One inequality rather than a guarantee
plus a caveat is what makes the specification part of the guarantee. Three consequences follow, in the order
an operator would use them. \emph{The direction matters, not the rate:} $\rpf$ enters only through
$\underline\kappa$ and degrades \Cref{eq:floorbias} smoothly for every $\rho < \tfrac12$, while $\rfp$ enters
additively and at $\bar\rho_F \ge \pt$ makes eviction inoperative discontinuously, so a judge with a $40\%$
error rate all in the harmless direction is safer than one with $46\%$ in the harmful one. \emph{Do not spend
the budget on $\Nmin$:} more trials shrink $\epsilon$ and cannot touch $\bar\rho_F$. The only lever inside the
algorithm is $\tau$, and it runs against intuition, the edge sitting at $\bar\rho_F = (1-\tau)/2$: a
\emph{narrower} margin tolerates a worse judge, and narrowing it is bounded in turn by $\epsilon$, which is
what A4 measures (\Cref{sec:evidence-ablations}). \emph{Move the judge instead:} any mechanism trading false
passes for phantom failures buys the guarantee back, which is what makes calibrated
abstention~\citep{jung2024trustescalate} the natural remedy and \Cref{sec:reward} the place to read it off a
measured judge.

\section{Evidence under an honest reward}
\label{sec:evidence}

\Cref{prop:floor} bounds the downside and says nothing about the upside, so the upside has to be measured.
It is $+0.328$ held-out $\passk$ on a hard MBPP+ slice, from a fixed author under a varying controller, and
the eight conditions below trace it to four mechanisms, fewer than Ratchet implements.

\paragraph{Protocol.}
The benchmark is MBPP+~\citep{liu2023evalplus}, the augmented-test-suite version of
MBPP~\citep{austin2021mbpp}, from which we take a fixed $100$-task slice, written
\bmbpp{} below: $60$ train and $40$ held-out eval, keeping only tasks the frozen base model fails on at
least one probe seed, so the slice isolates instances where a library could plausibly help. The split is
fixed across conditions and skills come only from train failures. The frozen model in all five
roles is Claude Opus 4.7~\citep{anthropic2026opus47}; runs are 100 rounds over three seeds ($42$, $7$,
$13$). The headline statistic is the \emph{rolling gain}, mean of the last ten rounds minus mean of the
first ten of held-out $\passk$, computed within each run so round-0 variation cancels; we also report the
peak. The reward is a unit-test grader with a near-zero false-pass rate, so assumption (ii) holds by
construction, which is the reason \Cref{sec:reward} exists. \Cref{app:config} gives every parameter and
role prompt, \Cref{app:seeds} the per-seed value behind every mean below.

\begin{figure}[t]
  \centering
  \includegraphics[width=\textwidth]{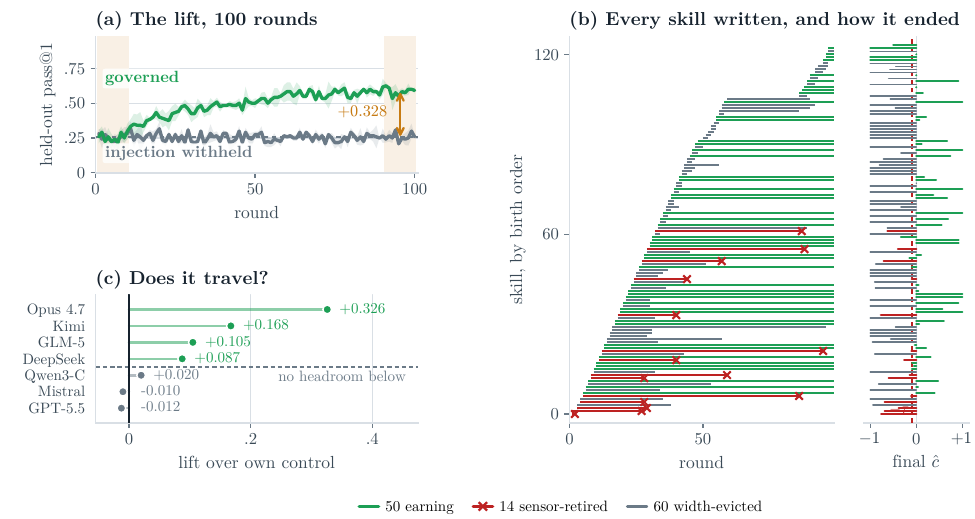}
  \caption{\textbf{Under an honest sensor: the lift, every skill's life, and whether both travel.}
  \emph{(a)} Held-out $\passk$ per round on \bmbpp{}, three seeds, $\pm 1\sigma$ ribbons. The closed loop
  (green) leaves the no-skill level and holds; withholding injection (grey) stays flat. Amber marks the two
  rolling-gain windows the $+0.328$ is computed from. \emph{(b)} The controller resolved to individual
  decisions: one row per skill ever written in seed 13, ordered by birth, drawn from the round it entered
  the shelf to the round it left, so the silhouette is the active count against the cap $C$. The right
  strip gives each skill's final $\chat$ against the eviction margin $-\tau$: survivors average $+0.29$,
  sensor-retired skills $-0.45$. \emph{(c)} Within-family lift, Default minus its own
  injection-withheld control, seven frozen models. Below the dashed rule the subset offers no headroom,
  having been filtered by Opus 4.7's failures (\Cref{sec:evidence-travel}).}
  \label{fig:curves}
\end{figure}

\paragraph{The lift, and the mechanism visibly at work.}
Held-out $\passk$ rises from $0.258 \pm 0.047$ at round 0 to a peak of $0.658 \pm 0.042$, a rolling gain of
$+0.328 \pm 0.018$, against $+0.002 \pm 0.005$ for the same loop with injection withheld. The three per-seed
gains fall within $0.04$ of each other (\Cref{tab:app-seeds}), which licenses attributing the lift to the
machinery rather than to one fortunate trajectory. Since the author is fixed and only the controller varies,
this is the thesis in its most direct measurable form: held-out $\passk$ more than doubles on tasks the same
model fails. \Cref{fig:curves} adds two things the scalar cannot. The trajectory climbs and holds rather than
spiking, distinguishing accumulated competence from a lucky round. And resolving the shelf into its $124$
skills shows the controller doing what the bound assumes: growth and eviction balance at fixed width, and the
two removal paths behave as different objects, sensor retirements arriving late at a median $107$ trials while
width evictions remove young skills on very few. Only the first is the eviction lever, which is why
\Cref{sec:reward} tracks it separately.

\subsection{Eight single-knob conditions}
\label{sec:evidence-ablations}

Each condition A1 to A8 changes exactly one knob and holds the rest fixed, so its rolling gain measures
that knob (\Cref{tab:ablations}). Forcing the Router to decline (A1) removes the entire gain while the loop
keeps synthesising and curating normally, so the value is in \emph{using} the shelf rather than maintaining
it, which also makes A1 the control \Cref{eq:harm} is defined against. Replacing the routing decision with
the top retrieval hit (A2) recovers under a quarter of the gain, so adjudicating which skill applies
contributes well beyond embedding similarity.

\emph{Breaks it (A4), the most useful finding here.} Tightening eviction to fire early and aggressively,
$\Nmin$ lowered to $20$ and $\tau$ to $0$, does not reduce the gain; it drives it to $-0.019 \pm 0.010$,
below the no-skill control. Governance done badly is worse than none, by the mechanism \Cref{sec:theory}
predicts: at $\Nmin = 20$ the Hoeffding radius grows to $\epsilon \approx 0.44$, and with $\tau = 0$ that
admits eviction of a strongly positive skill on unlucky early draws (\Cref{app:proofs}). The
shelf collapses to two active skills, and both survivors hurt. With A1 this brackets \Cref{eq:harm} from
both sides: withholding injection is inert, evicting without an evidence floor is actively harmful. It also
prices the one lever \Cref{sec:theory-bias} leaves for a lenient judge, since narrowing $\tau$ is what
raises the tolerable false-pass rate and A4 is what that costs when $\epsilon$ is not paid down first.

The remaining four land within about $0.04$ of the Default, inside $\pm 2\sigma$ at three seeds, so we claim
no improvement from them; the conservative reading of A5 and A6 is that explicit deduplication is
\emph{unnecessary at this scale}, the authoring prior already enforcing enough stylistic homogeneity to make
it redundant, and we expect it to matter on larger, more heterogeneous suites. Four mechanisms are therefore
load-bearing, injection, the routing decision, the authoring prior, and eviction \emph{with} an evidence
floor, so the minimal working recipe is smaller than the full system implies.

\begin{table}[t]
  \centering
  \footnotesize
  \caption{\textbf{Eight single-knob conditions} on \bmbpp{}, mean $\pm$ std over three seeds,
  against the Default's $+0.328 \pm 0.018$, grouped by what the row settles rather than by index. The red
  row is the only setting landing below the no-skill control; the grey block sits inside $\pm 2\sigma$ at
  $n=3$, so we make the weaker claim there.}
  \label{tab:ablations}
  \setlength{\tabcolsep}{5pt}
  \renewcommand{\arraystretch}{1.0}
  \begin{tabular}{@{}l l c c l@{}}
    \toprule
    & Knob changed & Rolling gain & $\Delta$ & What the row settles \\
    \midrule
    \multicolumn{5}{@{}l}{\emph{\textcolor{vGreen}{The lift needs these}}} \\
    A1 & injection withheld        & $+0.002 \pm 0.005$ & $-0.326$ & using the shelf, not keeping it, is the value \\
    A2 & retrieval replaces router & $+0.077 \pm 0.065$ & $-0.251$ & adjudication beats similarity by $3{\times}$ \\
    A3 & authoring prior removed   & $+0.187 \pm 0.036$ & $-0.141$ & costliest of the three levers to drop \\
    \addlinespace
    \multicolumn{5}{@{}l}{\emph{\textcolor{vRed}{Governance can be worse than none}}} \\
    \rowcolor{vRedF}
    A4 & $\Nmin{\to}20$, $\tau{\to}0$ & $-0.019 \pm 0.010$ & $-0.347$ & no evidence floor $\Rightarrow$ below the control \\
    \addlinespace
    \multicolumn{5}{@{}l}{\emph{\textcolor{vMuted}{Not necessary at this scale (differences inside $\pm 2\sigma$)}}} \\
    A7 & width $C{=}100$           & $+0.317 \pm 0.110$ & $-0.011$ & width non-binding in mean, $6{\times}$ variance \\
    A6 & cover-guard removed       & $+0.363 \pm 0.033$ & $+0.035$ & subsumed by the authoring prior \\
    A8 & prior refreshed / 10 rds  & $+0.372 \pm 0.017$ & $+0.044$ & best peak, $55\%$ more wall time \\
    A5 & canonicalisation removed  & $+0.374 \pm 0.023$ & $+0.046$ & subsumed by the authoring prior \\
    \bottomrule
  \end{tabular}
\end{table}

\subsection{Does it travel? Seven models and a different solver}
\label{sec:evidence-travel}

Every number so far holds the frozen model at Opus 4.7, which isolates the controller but leaves open
whether the lift belongs to the lifecycle or to that one model. We rerun the identical protocol, same
subset, same 100 rounds, same $\tau/\Nmin/C$, both the Default and its injection-withheld control, changing
only the model id, across six further families spanning six vendors and both open and closed weights: Kimi
K2.5~\citep{moonshot2026kimi}, GLM-5~\citep{zai2026glm5}, DeepSeek V3.2~\citep{deepseek2026v32},
Qwen3-Coder 480B~\citep{alibaba2026qwen3coder}, Mistral Large 3~\citep{mistral2026large3}, and
GPT-5.5~\citep{openai2026gpt55}. These runs are single-seed ($s{=}42$), so we read them as a generality
signal and report all six rather than the three that separate. The signal is the within-family gap, Default
minus its own control, which charges any lift to the controller rather than to raw capability.

That gap is clearly positive for three families, $+0.168$ (Kimi), $+0.105$ (GLM-5), and $+0.087$
(DeepSeek), with Kimi tracing the same monotone climb as Opus (\Cref{fig:curves}c). It is smaller than
Opus's $+0.326$ for the same reason the remaining three do not register: \bmbpp{} was filtered by
\emph{Opus's} failures, so every other model faces a subset selected against a stronger model. Qwen3-Coder
and Mistral Large 3 stay flat at $\leq +0.02$ because these tasks are largely out of reach, and no skill
teaches a model to solve what it fundamentally cannot; GPT-5.5 fails for the opposite reason, reaching a
$0.350$ peak on a first-ten window already elevated at $0.212$. The rolling gain therefore reports a lift
only in a middle capability band, and the three non-reproducing families are excluded by headroom or by the
statistic, not by the mechanism (\Cref{tab:app-crossmodel}).

To check that the levers act on the log rather than on a solver shape, we run the full Default loop on a
$150$-instance slice of SWE-bench Verified~\citep{jimenez2023swebench}, written \bswe{}, with skills injected
as a startup preamble. The single-call Solver is replaced by an agentic session that browses files, runs
tests, and iterates, in the style of SWE-agent~\citep{yang2024sweagent} and executable-action
agents~\citep{wang2024codeact} rather than the pipeline structure of Agentless~\citep{xia2024agentless}.
Across three seeds the no-skill agent averages $0.65$ held-out $\passk$ and injection lifts the mean peak
to $0.87$. We label this preliminary and report the peak only, since under twenty rounds, at roughly $50$
minutes each, leaves no stable late window (\Cref{app:crossmodel}).

\paragraph{Two threats to this reading.}
Rollback reads the same held-out signal we report, and three seeds is too few for hypothesis testing.
Neither survives inspection (\Cref{app:threats}): the guard fired \textbf{twice in 300 round-decisions}, so
the $+0.328$ holds with the net essentially off, and every separation we claim is an order of magnitude
larger than the per-seed spread. The load-bearing limitation is instead that $\chat$ is a decision statistic
and not a treatment effect (\Cref{sec:close}).

\section{The sensor specification, measured}
\label{sec:reward}

Everything in \Cref{sec:evidence} was graded by unit tests, the regime in which assumption (ii) is free.
The deployments expanding fastest, deep research and long-form report composition, are reference-free by
construction: no golden answer exists, so the only scalable reward is an LLM
judge~\citep{zheng2023llmjudge}, a choice already load-bearing in deployed systems such as
SkillForge~\citep{liu2026skillforge}. So the specification has to be measured, not assumed, and once rather
than never is the floor on that work: judge stability under prompt variation is itself an open measurement
problem~\citep{choi2026irtjudge}, and an optimiser pushing on a judge finds its blind spots rather than the
task~\citep{helff2026gamingverifiers}, which is the coupled regime \Cref{sec:close} excludes. This section
supplies the offline instrument, then tests both edges of \Cref{sec:theory-bias} in a running loop.

\subsection{Measuring the rate before deployment}
\label{sec:reward-gate}

The dangerous quantity $\rfp$ is estimable without running any evolution, which is what turns
\Cref{prop:floorbias} into an instrument. The construction needs a task that is reference-free yet carries an
objective sub-signal usable as ground truth, and citation-grounded report composition qualifies: no gold
report exists, but citation discipline is mechanically checkable. We take a deterministic grader of five
checks, a lightweight instance of the compact executable verifiers of
\citet{pezeshkpour2026autopyverifier}, as ground truth and inject defects into clean sections one at a time.
Five of seven classes mirror the checks and are \emph{check-visible}; two corrupt semantics while leaving
every annotation well formed, so no such check can see them. Over $155$ sections the grader catches every
check-visible injection and essentially none of the rest. A held-out judge from a different model family,
blind to condition, is complementary in exactly that gap: it flags the two semantic classes at $92.7\%$ and
$98.5\%$ and does not move on the structural ones. The audit is their union (\Cref{app:gate}). Applied to a
strict, well-instructed judge it measures $\rfp \approx 0.01$ and $\rpf \approx 0.95$: not the dangerous
corner but the conservative one, clearing the false-pass threshold by a factor of $45$ while sitting well past
the starvation edge $\rpf = 1 - \pt$, where the theory predicts safety bought with a channel resolution of
$\kappa \approx 0.04$ that barely separates good skills from bad. The lenient region is easy to enter from
here, a softer rubric, a fluent composer, or any defect class the judge cannot see moving $\rfp$ upward
without announcing it.

\subsection{Does the predicted boundary appear?}
\label{sec:reward-frontier}

\Cref{sec:theory-bias} predicts something of an unusual shape: not that higher error is worse, but that one
direction degrades smoothly for every rate below $\tfrac12$ while the other holds and then fails abruptly at
$\pt$. We test it by running the governed loop under a sweep of injected channels on \brmain{}, a $71$-prompt
slice of a reference-free long-form report-composition testbed. The bias arm then repeats on three further
slices that vary how often the composer fails: \brband{}, \brhard{}, and \bmbpp{} (Claude Haiku
4.5~\citep{anthropic2025haiku45}, three seeds, $\tau = 0.10$ and so the same $\pt = 0.45$, at a smaller
lifecycle budget to make a two-dimensional sweep affordable; \Cref{app:config}). Corruption is injected on
the \emph{training} reward only while evaluation uses the true grader, so movement in eval reflects a
reshaped library rather than a corrupted measurement of it.

\begin{figure}[t]
  \centering
  \includegraphics[width=\textwidth]{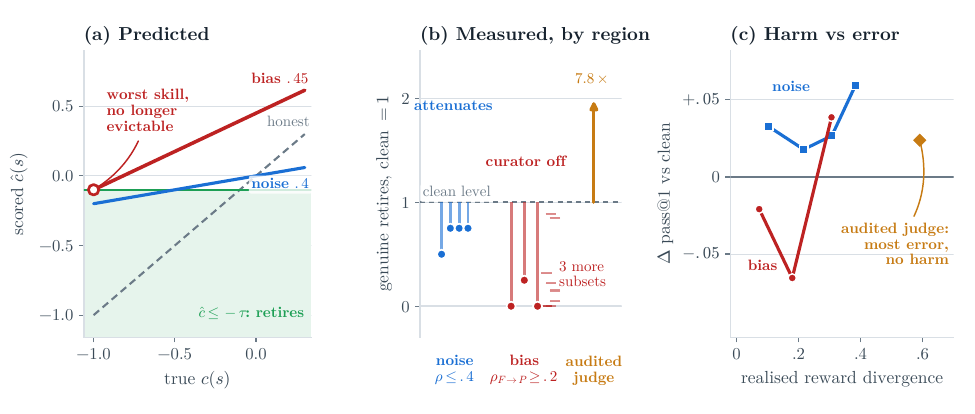}
  \caption{\textbf{The predicted boundary, then the same boundary measured.}
  Three seeds throughout. \emph{(a)} The prediction of
  \Cref{sec:theory-bias}, scored against true contribution: noise (blue) contracts about the origin so the
  crossing with $-\tau$ survives, bias (red) translates it upward, and at $\pt$ the crossing has left the
  domain, the hollow marker being the worst possible skill, $c = -1$. \emph{(b)} Genuine
  contribution-eviction per run, separated from width churn and normalised to the clean loop, each condition
  drawn as its drop from that level and grouped by region of \Cref{fig:thesis}b rather than by rate. Bold is
  the \brmain{} sweep, pale ticks the bias arm on \brband{}, \brhard{}, and \bmbpp{}. \emph{(c)} Held-out
  $\passk$ against the clean loop on \brmain{}, plotted against error delivered rather than rate injected.}
  \label{fig:frontier}
\end{figure}

\paragraph{Mechanism.}
The quantity to read is \emph{genuine} contribution-based eviction, since the raw deprecation count mixes it
with width churn, stays near ten under both corruptions, and hides the effect. Separated, the asymmetry is
sharp (\Cref{fig:frontier}b). Bias drives genuine eviction to zero or nearly zero at every injected rate and
starves synthesis too, skills written falling from $22$ to $15$, while noise only attenuates, holding
between $0.67$ and $1.00$ against $1.33$ clean (\Cref{app:sweep}). The audited judge fails from the far
side, retiring $10.3$ skills per run while carrying the largest realised divergence in the sweep, $0.59$,
essentially all phantom failures: highest total error, harmless direction, mechanism alive. The danger axis
is direction, not rate.

One condition qualifies the rectangle rather than the asymmetry. At $\rfp = 0.20$ the loop lies
\emph{inside} it and still retires nothing, $\pstar$ having fallen below the true pass rates of the skills
that shelf holds. The region is therefore necessary and not sufficient: it says when the rule can inform,
not that a population leaves it anything to fire on.

\paragraph{Outcome: end-task score is the wrong alarm.}
Eval quality tracks the mechanism (\Cref{fig:frontier}c). Symmetric noise never hurts at any rate tested,
sitting at or above the clean loop; the margins are inside the spread we decline to claim on elsewhere, so we
read them as the absence of harm rather than as a benefit and offer no mechanism for the sign. False-pass bias
is worst \emph{at} the boundary, $-0.065$, and by $\rfp = 0.7$ synthesis starves too and the nearly inert
shelf recovers past the clean level, $+0.039$, having too few surviving failures to reshape anything. That is
the non-monotonicity \Cref{sec:intro} promised, measured, and it disqualifies the dashboard three times over:
the worst excursion is a fifth of the $+0.328$ governance is worth, it is not ordered by the rate, and
eviction is deadest where it is smallest. The dead controller also travels while its visible cost does not.
Genuine eviction falls to essentially zero past the boundary on all three further subsets
(\Cref{fig:frontier}b), so a disabled controller is a property of the arithmetic rather than of a domain,
while the downstream cost appears only on the two failure-scarce slices, $-0.065$ here and $-0.036$ on
\brband{}, and not on \brhard{} or \bmbpp{}, where failures stay abundant enough to keep synthesis fed. A
deployment can be past the boundary and show nothing, which is the case the offline audit exists for.

\paragraph{Why the second edge costs resolution and not score.}
Indiscriminate eviction was catastrophic in \Cref{sec:evidence}, A4 landing at $-0.019$, and it is free
here, the audited judge reaching $+0.024$. The two are different failures wearing one description. A4
removed the evidence floor and so discarded skills whose true contribution was strongly positive,
collapsing the shelf to two survivors. The audited judge keeps the floor and is displaced only in the
harmless direction, so what it retires is disproportionately weak even though it retires far too much, and
synthesis refills the shelf against the cap. \Cref{prop:floorbias} says this is the general case and not the
lucky one, $\rpf$ entering as a divisor rather than additively, so past $1 - \pt$ the guarantee degrades
without breaking. What is lost is resolution, which an end-task score cannot see, and that is why the audit
reads both rates.

\paragraph{The floor holds where eviction does not.}
Consistent with \Cref{prop:floorbias}, the non-divergence that bounded width promises held across the entire
sweep: all means stayed within $0.125$ of the no-skill control, including where eviction was inoperative.
That is the division of labour \Cref{sec:loop} derived the two levers to have. Code with unit tests is safe
not because failures are plentiful there but because its grader is a sound verifier, and \bmbpp{} collapses
like any other domain once a false-pass channel is injected on it. So measure both rates offline. Inside the
rectangle eviction works; past $\pt$, lower the effective $\rfp$ with a sharper rubric, a held-out judge, or
abstention on low-confidence passes, and past $1 - \pt$ soften the rubric instead, or leave the library
unmaintained until the sensor can carry it.

\section{Scope and limitations}
\label{sec:close}

\paragraph{The result is about any threshold-retired artifact store.}
Neither proposition mentions a skill. Both hold for any store retired by a threshold rule on one scalar per
artifact, so a memory bank, a retrieval corpus, or a score-pruned tool registry inherits the same floor, the
same $\pt$, and the same obligation to audit its sensor first.

\paragraph{Limitations, in descending order of how much they constrain the conclusions.}
\emph{(1) The corruption is exogenous.} Injecting a channel on top of a deterministic grader is what isolates
it, so the coupled regime, where an evolving library learns phrasing that induces the blindness and $\rfp$
climbs along the trajectory, is out of scope and no fixed audit bounds it. That regime is not hypothetical:
\citet{helff2026gamingverifiers} show optimisation against an imperfect verifier locating the verifier's blind
spots rather than the task, which is the same mechanism with a policy in place of a library. We would attack it
first, the trajectory being recoverable post hoc from the same log, and a repeated audit across rounds turning
$\rfp$ from a constant into a series. \emph{(2) The audit certifies only enumerable
classes,} locating the boundary against failure modes an operator can name, not all of them.
\emph{(3) The sensor is associational.} $\chat$ carries a survivorship confound and a selection one, the
Router steering each skill toward tasks it already looks good on; \Cref{prop:floor} is stated for that same
routed-conditional target, so the floor holds for what the Curator reads, but a causal reading needs
randomised toggling. \emph{(4) Seeds.} Three support the load-bearing separations and not the $\sim 0.04$
differences we decline to claim, and the six further families are single-seed; since the floor rises with
base capability, governance complements scale rather than substituting for it~\citep{sutton2019bitter}.
\emph{(5) No governance bake-off,} the sharpest test being to swap each system's curator onto a shared log,
which the role-factored design supports.

\paragraph{Conclusion.}
Self-evolving libraries drift for want of maintenance, not for want of authorship, and maintenance is worth
exactly what its sensor is worth. What the derivation adds is that the worth is not continuous in sensor
quality: it has an edge at $(1-\tau)/2$, direction is the only thing that decides which side of it a judge
falls on, and past it no sample size and nothing on an end-task dashboard recovers the rule. So calibrate
the sensor first. The rate that disarms it is measurable in one offline pass before any evolution is run,
and nothing measures it afterwards.

\phantomsection\addcontentsline{toc}{section}{Broader impact}
\subsubsection*{Broader impact}
The result most consequential for practice is negative: lifecycle governance can stop working while the
end-task scores an operator watches barely move. We would rather that failure mode, and the one-pass
measurement for it, be known than met in a deployment. The matching risk is reading the measurement as
broader than it is, since it certifies a judge only against enumerable defect
classes (\Cref{sec:reward,sec:close}).

\bibliographystyle{plainnat}
\phantomsection

\appendix
\section{Per-seed numbers behind every mean in the body}
\label{app:seeds}

The body reports means with spreads. \Cref{tab:app-seeds} gives the constituent per-seed values so that
every aggregate can be recomputed, and so that the two claims we decline to make, on differences inside
$\pm 2\sigma$, can be checked rather than taken on trust. Rolling gain is the within-run window difference
defined in \Cref{sec:evidence}, so each cell below is one run's own statistic and the means are means of
these three columns.

\begin{table}[ht]
  \centering
  \footnotesize
  \caption{Rolling gain per seed, all nine conditions on \bmbpp{}, $100$ rounds each. The Default row is
  the headline $+0.328 \pm 0.018$. A7's spread is the widest in the table, which is the basis for reading
  its mean as non-informative rather than as a null result.}
  \label{tab:app-seeds}
  \setlength{\tabcolsep}{7pt}
  \begin{tabular}{@{}l l c c c c@{}}
    \toprule
    & Condition & $s{=}42$ & $s{=}7$ & $s{=}13$ & mean $\pm$ std \\
    \midrule
    -- & Default                     & $+0.302$ & $+0.340$ & $+0.343$ & $+0.328 \pm 0.018$ \\
    \addlinespace
    A1 & injection withheld          & $+0.003$ & $-0.005$ & $+0.008$ & $+0.002 \pm 0.005$ \\
    A2 & retrieval replaces router   & $+0.147$ & $+0.095$ & $-0.010$ & $+0.077 \pm 0.065$ \\
    A3 & authoring prior removed     & $+0.148$ & $+0.177$ & $+0.235$ & $+0.187 \pm 0.036$ \\
    A4 & $\Nmin{\to}20$, $\tau{\to}0$ & $-0.005$ & $-0.027$ & $-0.025$ & $-0.019 \pm 0.010$ \\
    A5 & canonicalisation removed    & $+0.343$ & $+0.385$ & $+0.395$ & $+0.374 \pm 0.023$ \\
    A6 & cover-guard removed         & $+0.365$ & $+0.403$ & $+0.322$ & $+0.363 \pm 0.033$ \\
    A7 & width $C{=}100$             & $+0.172$ & $+0.340$ & $+0.440$ & $+0.317 \pm 0.110$ \\
    A8 & prior refreshed / 10 rounds & $+0.365$ & $+0.355$ & $+0.395$ & $+0.372 \pm 0.017$ \\
    \bottomrule
  \end{tabular}
\end{table}

Round-0 held-out $\passk$ is $0.225$, $0.225$, and $0.325$ for seeds $42$, $7$, and $13$, mean
$0.258 \pm 0.047$; the per-seed peaks are $0.675$, $0.600$, and $0.700$, mean $0.658 \pm 0.042$. The
rollback net fired on zero, one, and one round-decision respectively, two firings in $300$.

\section{The cross-model probe, both halves}
\label{app:crossmodel}

\Cref{fig:curves}c plots the within-family gap only. \Cref{tab:app-crossmodel} gives both halves, which is
what makes the headroom argument checkable: the three families that do not separate have round-0 rates at or
below $0.125$ on a slice filtered by a stronger model's failures, so there is little for a skill to recover.

\begin{table}[ht]
  \centering
  \footnotesize
  \caption{Seven frozen models on \bmbpp{}, identical protocol, single seed ($s{=}42$) for the six
  non-Opus families. The gap is Default minus that family's own injection-withheld control, which charges
  any lift to the controller rather than to raw capability.}
  \label{tab:app-crossmodel}
  \setlength{\tabcolsep}{7pt}
  \begin{tabular}{@{}l l c c c c@{}}
    \toprule
    Model & Vendor & round-0 & Default & control & gap \\
    \midrule
    Claude Opus 4.7   & Anthropic & $0.258$ & $+0.328$ & $+0.002$ & $+0.326$ \\
    Kimi K2.5         & Moonshot  & $0.050$ & $+0.183$ & $+0.015$ & $+0.168$ \\
    GLM-5             & Z.ai      & $0.025$ & $+0.117$ & $+0.012$ & $+0.105$ \\
    DeepSeek V3.2     & DeepSeek  & $0.025$ & $+0.075$ & $-0.012$ & $+0.087$ \\
    \addlinespace
    Qwen3-Coder 480B  & Alibaba   & $0.025$ & $+0.020$ & $+0.000$ & $+0.020$ \\
    Mistral Large 3   & Mistral   & $0.025$ & $-0.010$ & $+0.000$ & $-0.010$ \\
    GPT-5.5           & OpenAI    & $0.125$ & $+0.003$ & $+0.015$ & $-0.012$ \\
    \bottomrule
  \end{tabular}
\end{table}

GPT-5.5 is the one case where the statistic rather than the mechanism is at fault: its peak reaches $0.350$
but its first-ten window is already elevated at $0.212$, so a difference of windows registers nothing. On
\bswe{} the three seeds run $18$, $18$, and $19$ rounds before the wall-clock budget of roughly $50$ minutes
per round exhausts, with round-0 held-out $\passk$ of $0.600$, $0.667$, and $0.683$ and peaks of $0.850$,
$0.917$, and $0.850$. We report the peak and label the result preliminary because at under twenty rounds no
late window is stable.

\section{The defect-injection audit, in full}
\label{app:gate}

The audit of \Cref{sec:reward-gate} estimates $\rfp$ and $\rpf$ for a candidate judge before any evolution is
run. We give its construction and runs in the detail needed to check the placement in \Cref{fig:thesis}b,
which reads both rates. It has two halves, and the second exists because the first is blind in a specific way.

\paragraph{Seven defect classes.}
Five mirror a deterministic check and are \emph{check-visible}: orphan citation (a marker with no
bibliography entry), unregistered metric (a quantity absent from the run's metric registry), unsourced
number (a bare figure with no citation on the line), broken cross-reference, and summary removal. The two
\Cref{sec:reward-gate} calls semantic are \emph{claim negation}, which flips the direction of a cited claim,
and \emph{number swap}, which perturbs a digit while keeping the citation marker in place. One defect is injected per class per section, over $155$ sections
($137$ for claim negation, where not every section carries a directional claim).

\paragraph{The deterministic grader is exactly as blind as designed.}
Recall is $1.00$ on all five check-visible classes ($n=155$ each). It is $0.00$ on claim negation and
$0.05$ on number swap. A loop rewarded by this grader alone therefore scores a semantically corrupted
section as a pass, which is a false pass in the sense of \Cref{eq:channel}.

\paragraph{The held-out judge is complementary in that gap.}
Because this half costs a judge call per section rather than a deterministic check, it runs on a subsample
rather than on all $155$: $143$ clean sections and their injected counterparts, with the two semantically
corrupted classes sampled most heavily since they are the ones the grader cannot see, and $44$ calls
discarded on API error. \Cref{tab:app-gate} gives the resulting pair count, flag rate, and mean drop in the
judge's scalar quality score per class. It flags
number swap at $0.985$ and claim negation at $0.927$, the two the grader misses, while on the two purely
structural classes its quality score does not move ($p \approx 0.5$), which is the sense in which the two
instruments are complementary rather than redundant. The audit is their union, and $\rfp$ is estimated as
the fraction of injected sections the union still scores as clean.

\begin{table}[ht]
  \centering
  \footnotesize
  \caption{The held-out judge on paired clean and injected sections. Mean drop is in its overall quality
  score; $p$ is a permutation test on that drop. The two rows in bold are the classes the deterministic
  grader cannot see, and are the two where the judge moves most.}
  \label{tab:app-gate}
  \setlength{\tabcolsep}{7pt}
  \begin{tabular}{@{}l c c c c@{}}
    \toprule
    Injected class & pairs & flag rate & mean drop & $p$ \\
    \midrule
    \textbf{number swap}          & $67$ & $\mathbf{0.985}$ & $1.194$ & $<10^{-4}$ \\
    \textbf{claim negation}       & $55$ & $\mathbf{0.927}$ & $0.255$ & $0.004$ \\
    \addlinespace
    unsourced number             & $27$ & $1.000$ & $2.074$ & $<10^{-4}$ \\
    unregistered metric          & $28$ & $1.000$ & $1.464$ & $<10^{-4}$ \\
    orphan citation              & $28$ & $1.000$ & $0.893$ & $<10^{-4}$ \\
    \addlinespace
    broken cross-reference       & $25$ & $0.840$ & $0.040$ & $0.500$ \\
    summary removal              & $26$ & $0.808$ & $0.038$ & $0.504$ \\
    \bottomrule
  \end{tabular}
\end{table}

Applied to the strict, well-instructed training judge, the union measures $\rfp \approx 0.01$ over $210$
injected check-visible sections and $\rpf \approx 0.95$ over the $42$ audited sections that truly pass, hence
$\kappa \approx 0.04$. That is a safe but resolution-poor sensor: it almost
never hides a failure, and it invents a great many, which is why \Cref{sec:reward} reads it as sitting
past the starvation edge $\rpf = 1 - \pt$ of \Cref{fig:thesis}b rather than inside the governable
rectangle. The two estimates are not equally tight, and the placement rests on the looser one: at $n=42$
the exact binomial $95\%$ interval on $\rpf$ spans $0.84$ to $0.99$, which lies wholly above the $0.55$
edge, so the side of the boundary is determined even though the rate is not pinned down.

\section{The corruption sweep, condition by condition}
\label{app:sweep}

\Cref{tab:app-sweep} gives every point behind \Cref{fig:frontier}b and c on \brmain{}, three seeds each,
at $\tau = 0.10$, $\Nmin = 24$, $C = 12$, organised by where each condition falls on the certification
plane rather than by its error rate. The organising column is $\pstar = (\pt - \rfp)/\kappa$, the largest
true pass rate a skill can have and still be evictable (\Cref{sec:theory-bias}): it is the single number
the two boundaries are statements about, and reading the rows against it accounts for the two retirement
counts the rates alone do not explain. Genuine retirement counts are per run
and separate contribution-driven retirement from width eviction; the raw deprecation count, which mixes the
two, stays near ten in every condition and is why the effect is invisible without the separation.

\begin{table}[ht]
  \centering
  \footnotesize
  \caption{Every sweep condition, keyed to the two boundaries. $\pstar = (\pt - \rfp)/\kappa$ is the
  eviction ceiling on true pass rate; it must lie in $(0,1)$ for the rule to inform, and the blocks are the
  three ways that can go. Divergence is the error the channel actually delivered, not the rate injected.
  $\Delta$ is held-out $\passk$ against the clean loop's $0.569$. Bold marks the two conditions whose
  retirement count $\pstar$ explains and the rates do not.}
  \label{tab:app-sweep}
  \setlength{\tabcolsep}{5.5pt}
  \begin{tabular}{@{}l c c c c c c@{}}
    \toprule
    Condition & $(\rfp, \rpf)$ & $\pstar$ & divergence & retires/run & written & $\Delta$ \\
    \midrule
    \multicolumn{7}{@{}l}{\emph{\textcolor{vGreen}{inside the rectangle: $\pstar \in (0,1)$, the rule
      informs}}} \\
    clean grader          & $(0, 0)$          & $0.45$ & $0.000$ & $1.333$ & $22.0$ & $\pm 0.000$ \\
    $\rho = 0.10$         & $(0.10, 0.10)$    & $0.44$ & $0.103$ & $0.667$ & $22.0$ & $+0.033$ \\
    $\rho = 0.20$         & $(0.20, 0.20)$    & $0.42$ & $0.216$ & $1.000$ & $23.0$ & $+0.018$ \\
    $\rho = 0.30$         & $(0.30, 0.30)$    & $0.38$ & $0.308$ & $1.000$ & $23.0$ & $+0.027$ \\
    $\rho = 0.40$         & $(0.40, 0.40)$    & $0.25$ & $0.384$ & $1.000$ & $24.0$ & $+0.060$ \\
    $\rfp = 0.20$         & $(0.20, 0)$       & $\mathbf{0.31}$ & $0.073$ & $\mathbf{0.000}$ & $19.7$
      & $-0.021$ \\
    \addlinespace
    \multicolumn{7}{@{}l}{\emph{\textcolor{vRed}{past the false-pass edge: $\pstar \le 0$, nothing is
      evictable}}} \\
    $\rfp = 0.45 = \pt$   & $(0.45, 0)$       & $0.00$ & $0.180$ & $0.333$ & $15.7$ & $-0.065$ \\
    $\rfp = 0.70$         & $(0.70, 0)$       & $< 0$  & $0.306$ & $0.000$ & $15.0$ & $+0.039$ \\
    \addlinespace
    \multicolumn{7}{@{}l}{\emph{\textcolor{vAmber}{past the starvation edge: $\pstar \ge 1$, everything
      is}}} \\
    audited judge         & $(0.01, 0.95)$    & $\mathbf{> 1}$ & $0.591$ & $\mathbf{10.333}$ & $19.0$
      & $+0.024$ \\
    \bottomrule
  \end{tabular}
\end{table}

Read down the $\pstar$ column, the table says three things the rates cannot. \emph{The ceiling is what
degrades, and noise barely moves it.} Symmetric error shrinks $\pstar$ from $0.45$ to $0.25$ across the
whole sweep and retirement survives throughout, because the rule still selects on the right side of the
population. \emph{The two bold rows are the test.} At $\rfp = 0.20$ the condition is inside the rectangle
yet retires nothing: $\pstar$ has fallen to $0.31$, below the true pass rate of the skills this shelf
actually holds. That is what makes the rectangle necessary and not sufficient, and no reading of the rate
predicts it. At the other edge the audited judge has $\pstar > 1$ and retires $10.3$ skills per
run against $1.3$ clean, the same rule firing on everything, and it does so while delivering the largest
divergence in the sweep, $0.591$, almost all of it phantom failures. Highest total error, opposite edge,
opposite failure. At $\rfp = \pt$ exactly, $\pstar = 0$ and the surviving $0.333$ retirements per run are
the finite-sample residue the boundary is asymptotic about: $\bar{p}(s) \le 0$ is unsatisfiable in the mean,
not in every draw of $24$ trials. \emph{Harm is not monotone in either rate.} Held-out quality is worst at the false-pass
boundary, $-0.065$, and recovers to $+0.039$ past it, because at $\rfp = 0.70$ so few failures survive
scoring that synthesis starves: skills written falls from $22$ to $15$, which is that starvation measured
directly rather than inferred.

\section{The two threats to \texorpdfstring{\Cref{sec:evidence}}{the evidence section},
worked through}
\label{app:threats}

\paragraph{Rollback is not selecting the trajectory.}
The guard of \Cref{sec:loop} reads the same held-out signal the body reports, so a reader is right to ask
whether the reported trajectory is chosen rather than earned. Recomputed across the three seeds, it fired on
zero, one, and one round-decision: it arrested two sustained regressions and was otherwise inactive. Two
further points close the concern. The statistic is a within-run difference of
windows, so it is not a quantity round-selection can inflate: restoring a snapshot moves the late window and
the early window it is measured against alike. And we add no third split deliberately, because a deployed
self-improving agent rolls back against its live use cases, not against a validation proxy it does not
possess; adding one would measure a system nobody runs.

\paragraph{Three seeds, and what we therefore do not claim.}
Three seeds do not support hypothesis testing, so we report spread rather than intervals throughout and
keep the claims to separations large relative to it. The headline $+0.328 \pm 0.018$ clears the $+0.002$
injection-withheld control and the $-0.019$ harsh-eviction floor by more than an order of magnitude of the
per-seed spread, and \Cref{tab:app-seeds} shows every seed on the same side of both in every case. The
four conditions inside $\pm 2\sigma$ of the Default are the ones this resolution cannot separate, which is
why \Cref{sec:evidence-ablations} makes only the weaker claim about them, and A7 is the extreme case: a
spread of $\pm 0.110$ over per-seed values of $+0.172$, $+0.340$, and $+0.440$ is not a null result but an
uninformative one. The six additional model families are single-seed and read as a generality signal
rather than as measurements.

\section{Proofs}
\label{app:proofs}

\Cref{sec:theory} states both propositions and the one-line reason each holds. Both are proved in full here,
followed by the two steps worth expanding.

\paragraph{Proof of \Cref{prop:floor}.}
Work on the event of assumption (iii), which holds with probability at least $1 - C\delta$ by a union bound
over the at most $C$ members of $\mathcal{S}_t$. Any $s$ that survives to round $t$ failed the eviction
test, so $\chat(s) > -\tau$, and on the concentration event $c(s) \ge \chat(s) - \epsilon > -\tau -
\epsilon$. For tasks routed to such an $s$, \Cref{eq:contrib} gives
$\mathbb{E}_{\mathcal{D}_s}[p(x \mid s)] \ge \mathbb{E}_{\mathcal{D}_s}[p_0(x)] - \tau - \epsilon$, while
declined tasks attain $p_0(x)$ exactly by assumption (i). Averaging over the routing distribution, the
$\mathbb{E}_{\mathcal{D}_s}[p_0]$ terms recombine into $\mathbb{E}[p_0]$ and every routed group carries at
most an additive $\tau + \epsilon$ deficit, so the conditional expectation on the concentration event is at
least $\mathbb{E}[p_0] - (\tau + \epsilon)$. Charging the complementary event, of probability at most
$C\delta$, its worst case yields \Cref{eq:floor}. \hfill$\square$

\paragraph{Proof of \Cref{prop:floorbias}.}
On the concentration event, again of probability at least $1 - C\delta$, every surviving skill has scored
pass rate at least $\pt - \epsilon$, since it failed the test $\hat p(s) \le \pt$. \Cref{eq:pobs} is
$p_{\mathrm{sc}} = \kappa \bar p + \rfp$ with $\kappa > 0$, hence strictly increasing in $\bar p$ and
invertible, so $\bar p(s) = (p_{\mathrm{sc}}(s) - \rfp)/\kappa \ge (\pt - \epsilon - \rfp)/\kappa$, which
the assumed bounds weaken to $\bar p(s) \ge (\pt - \epsilon - \bar\rho_F)/\underline\kappa$. A task routed
to a survivor therefore has true pass probability at least this value, while a declined task attains
$p_0(x)$. Averaging as in \Cref{prop:floor}, now on the $[0,1]$ pass-rate scale where the margin is
$\tau/2$ and $\pt = \tfrac12 - \tau/2$, substituting and collecting the constant relative to $\tfrac12$
produces both the $(\tau/2 + \epsilon + \bar\rho_F)/\underline\kappa$ term and the residual
$\tfrac12(1 - \underline\kappa)$, and charging the complementary event $C\delta$ yields
\Cref{eq:floorbias}. Setting $\bar\rho_F = \bar\rho_P = 0$ gives $\underline\kappa = 1$ and recovers
\Cref{eq:floor}: the two propositions are one statement read at two sensor qualities. \hfill$\square$

\paragraph{The concentration event, and why $C$ enters linearly.}
Assumption (iii) of \Cref{prop:floor} asks that $|\chat(s) - c(s)| \le \epsilon$ hold simultaneously for
all $s \in \mathcal{S}_t$. Since $\chat = 2\hat p - 1$ takes values in $[-1,1]$, Hoeffding's inequality
gives $\Pr[|\chat(s) - c(s)| > \epsilon] \le 2\exp(-\Nmin \epsilon^2 / 2)$ for a single skill, so setting
that bound to $\delta$ yields $\epsilon = \sqrt{2\ln(2/\delta)/\Nmin}$. At $\Nmin = 100$ and
$\delta = 10^{-3}$ this is $\epsilon \approx 0.39$. A union bound over the at most $C$ active skills
charges $C\delta$ to the complementary event, and this is the only place the width enters the probability:
it is also why an unbounded shelf has no counterpart to \Cref{eq:floor}, since $C\delta$ then diverges
however small $\delta$ is made. At the A4 setting of $\Nmin = 20$ the same expression gives
$\epsilon \approx 0.44$, which with $\tau = 0$ admits eviction of a skill whose true contribution is as
high as $+0.44$, and \Cref{tab:app-seeds} shows that configuration falling below the no-skill control on
all three seeds.

\paragraph{Why the inversion is where the discontinuity comes from.}
The two corruption rates enter \Cref{eq:floorbias} through different routes, which is what makes the
specification two-sided rather than a single error budget. $\bar\rho_F$ enters the numerator additively, so
at $\bar\rho_F = \pt$ the numerator reaches $\pt$ and the bound on a survivor's true rate collapses to
$0$: the eviction test is then satisfiable by no skill, and no choice of $\Nmin$ recovers it, since
$\epsilon$ appears in the same numerator and shrinking it cannot offset a term that does not shrink.
$\bar\rho_P$ enters only through $\underline\kappa = 1 - \bar\rho_F - \bar\rho_P$ as a divisor, so it
inflates the deficit smoothly for every rate below $\tfrac12$ and the guarantee degrades without failing.
Additive versus divisor is the whole asymmetry, and it is why direction rather than total error rate decides
whether the controller keeps working.

\section{Configuration}
\label{app:config}

\paragraph{Defaults.}
$\tau = 0.10$, $\Nmin = 100$, $C = 50$, rollback threshold $\tau_{\mathrm{rb}} = 0.10$ with five
consecutive flags required, canonicalisation cosine threshold $0.85$, minimum cluster size $3$ for
synthesis. The corruption sweep uses $\tau = 0.10$ with the smaller lifecycle budget $\Nmin = 24$ and
$C = 12$ so that a two-dimensional sweep is affordable; $\pt = (1-\tau)/2 = 0.45$ is set by $\tau$ alone
and so is identical in both configurations.

\paragraph{Compute.}
Every cost is inference against a frozen served model; \Cref{sec:loop} gives the per-round call count. On that
budget the single-call \bmbpp{} rounds are cheap, while the agentic \bswe{} loop at roughly $50$ minutes per
round is what bounds those runs to under twenty rounds.

\paragraph{What each role is given.}
\Cref{sec:loop} names the five roles; this is the input each one receives, which is what a reimplementation
needs. The \textsc{Router} gets the task and the active shelf as (name, applicability condition) pairs, and
returns one skill id or a decline. The \textsc{Solver} gets the task with the routed skill's body prepended.
The \textsc{Critic} gets the failing capsule, its task, the routed skill, the output, and the grader trace,
and returns one label from $\{$\textsc{helped}, \textsc{hurt}, \textsc{neutral},
\textsc{inapplicable}$\}$ with a pattern string. The \textsc{Synthesizer} gets a cluster of pattern strings
and the meta-skill document. The \textsc{Curator} gets $\chat(s)$ and $n(s)$, and makes no model call.

\paragraph{Which conditions touch the authoring prior.}
A3 removes the meta-skill document entirely, A8 rewrites it every ten rounds from the accumulated log, and
A7 leaves it untouched while doubling the width to $C=100$.

\typeout{get arXiv to do 4 passes: Label(s) may have changed. Rerun}
\end{document}